\documentclass[10pt,twocolumn,letterpaper]{article}

\usepackage[pagenumbers]{cvpr} % To force page numbers, e.g. for an arXiv version

\usepackage{caption}

\usepackage[dvipsnames]{xcolor}

\usepackage{adjustbox}
\usepackage{multirow}
\usepackage[table]{xcolor}
\usepackage{bm}
\usepackage{booktabs}
\usepackage{makecell}
\usepackage{pifont}
\usepackage{lipsum}
\usepackage{wrapfig}
\newcommand{\repname}{Floorplan Markup Language\xspace}
\newcommand{\repnameshort}{FML\xspace}
\newcommand{\modelname}{\repname Model\xspace}
\newcommand{\modelnameshort}{{\repnameshort}M\xspace}

\definecolor{cvprblue}{rgb}{0.21,0.49,0.74}
\usepackage[pagebackref,breaklinks,colorlinks,allcolors=cvprblue]{hyperref}

\def\paperID{} % *** Enter the Paper ID here
\def\confName{CVPR}
\def\confYear{2026}

\title{Unified Vector Floorplan Generation via Markup Representation}

\author{\stepcounter{footnote}Kaede Shiohara
$\quad$$\quad$ Toshihiko Yamasaki \\ 
The University of Tokyo\\ 
{\tt\small \{shiohara, yamasaki\}@cvm.t.u-tokyo.ac.jp}
}

\begin{document}
\twocolumn[{%
    \renewcommand\twocolumn[1][]{#1}%
    \maketitle
    \begin{center}
      \includegraphics[width=\textwidth]{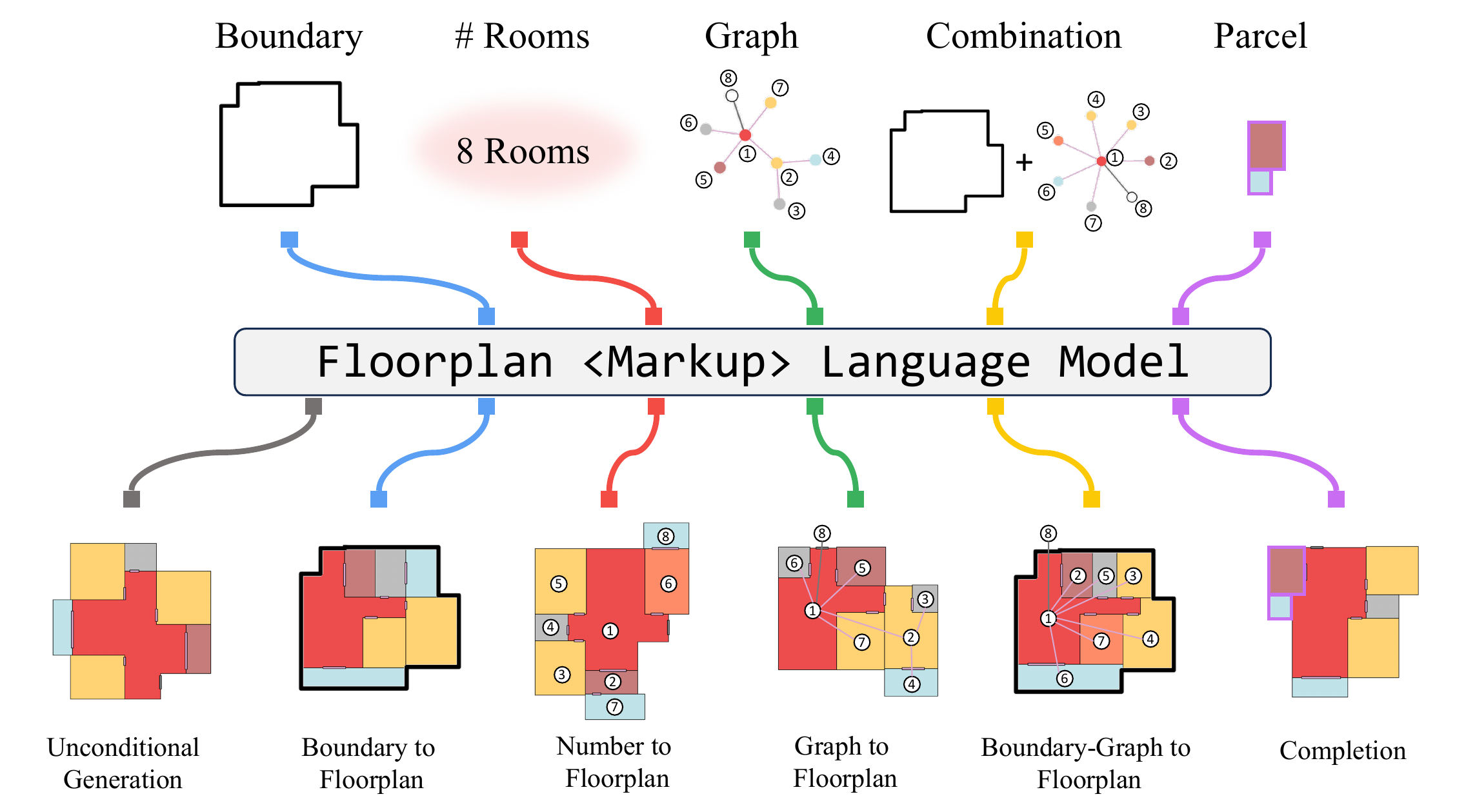}
      \captionof{figure}{Our \modelname directly generates vector floorplans under the wide range of situations. 
      }
      \label{fig:teaser}
    \end{center}
}]

\begin{abstract}
Automatic residential floorplan generation has long been a central challenge bridging architecture and computer graphics, aiming to make spatial design more efficient and accessible.
While early methods based on constraint satisfaction or combinatorial optimization ensure feasibility, they lack diversity and flexibility. 
Recent generative models achieve promising results but struggle to generalize across heterogeneous conditional tasks, such as generation from site boundaries, room adjacency graphs, or partial layouts, due to their suboptimal representations.
To address this gap, we introduce \repname (\repnameshort), a general 
representation that encodes floorplan information within a single structured grammar, which casts the entire floorplan generation problem into a next token prediction task.
Leveraging \repnameshort, we develop a transformer-based generative model, \modelnameshort, capable of producing high-fidelity and functional floorplans under diverse conditions.
Comprehensive experiments on the RPLAN dataset 
demonstrate that \modelnameshort, despite being a single model, surpasses the previous task-specific state-of-the-art methods. 
Project page: \url{https://mapooon.github.io/FMLPage}.
\end{abstract}
    
\section{Introduction}
\label{sec:intro}
Automatically generating residential floorplans has long been a goal at the intersection of architecture and computer graphics. 
It promises to lower design costs, accelerate early-stage exploration, and empower non-experts to test multiple layout options quickly. 
A key challenge is to generate layouts that satisfy practical design constraints while remaining plausible. 
This typically means honoring user-specified inputs such as site boundaries, room types, or adjacency graphs while producing functionally valid and architecturally realistic layouts.

Early systems~\cite{10.1007/s00371-013-0825-1,10.1111:cgf.13380,LAIGNEL2021103491} formulated floorplan generation as constraint satisfaction or combinatorial search, arranging rooms under adjacency, area, and circulation rules.
While these approaches provide hard feasibility guarantees, they require extensive expert-crafted heuristics and offer limited stylistic diversity and suffer from computational overheads for each single floorplan generation.
With the rapid progress of generative models~\cite{gan,vae,ddpm}, data-driven approaches have emerged. 
Some methods generate rasterized segmentation masks of room types~\cite{housegan,houseganpp} while others operate directly in vector space to synthesize polygonal layouts~\cite{housediffusion,cons2plan,gsdiff} empowered by diffusion models~\cite{ddpm}.
However, despite these efforts, previous approaches suffer from poor generalizability across different types of conditional generation tasks such as boundary, number of rooms, room adjacency graph, completion from partial layouts, and their combinations.
This is because previous methods rely on suboptimal representations that are less compatible with the structural information of floorplan data, which significantly makes the previous approaches inefficient and redundant by requiring conversion from raster to vector floorplans~\cite{housegan,houseganpp} and multi-stage pipelines~\cite{graph2plan,cons2plan,gsdiff}.

To address this issue, we propose a general representation called \repname (\repnameshort). 
Inspired by HyperText Markup Language (HTML), \repnameshort represents a floorplan and its conditions such as a boundary and graph as a single sequence of tagged elements, where the tag structure explicitly constrains which elements can appear next.
The structured, tag-based design of \repnameshort provides a clear and regular syntax, which naturally constrains the generation process and guides the model toward valid and coherent layouts.
Leveraging this representation, our auto-regressive transformer model trained to generate \repnameshort sequences, which is called the \modelname (\modelnameshort), can produce high-fidelity floorplans across a wide variety of tasks, as illustrated in Fig.~\ref{fig:teaser}.

We compare our model with the previous state-of-the-art methods such as Graph2Plan~\cite{graph2plan}, HouseGAN++~\cite{houseganpp}, HouseDiffusion~\cite{housediffusion}, and GSDiff~\cite{gsdiff} on the RPLAN dataset~\cite{rplan}.
Extensive experiments show that our unified model outperforms the previous task-specific models in a wide range of tasks in terms of FID~\cite{fid}, IoU, and GED~\cite{ged}.
\section{Related Work}
\label{sec:related_work}

\subsection{Floorplan Generation}
\noindent\textbf{Optimization-based approaches.}
Early work tried to generate floorplans by iterative optimization on hand-crafted rules and user-specified constraints such as property boundaries, size of rooms, and adjacencies between rooms.
Liu \etal~\cite{10.1007/s00371-013-0825-1} proposed an interactive pipeline with an active-set optimizer that minimizes the costs for area fidelity, aspect-ratio regularization, boundary utilization, and a fabrication term that aligns wall lengths to a catalog of precast slab widths.
IP-Layout~\cite{10.1111:cgf.13380} introduced a hierarchical, coarse-to-fine sub-domain refinement that scales to large layouts, \eg, offices, malls, and supermarkets, solved with a mixed integer quadratic programming solver.
Optimizer~\cite{LAIGNEL2021103491} integrated genetic optimization into constraint programming, which improves the functionality and architectural fidelity such as room shape alignment and circulation efficiency.

However, these methods suffer from the trade-off between fidelity and diversity; more rules decrease the diversity of floorplans while less rules degrade the fidelity.

\begin{table}[t]
\centering
\begin{adjustbox}{width=1.0\linewidth}
\begin{tabular}{c|ccccccc}
\toprule
\multirow{2}{*}{Method} & \multirow{2}{*}{Uncond.} &\multicolumn{4}{c}{Conditional} & \multirow{2}{*}{Compl.}\\
\cmidrule(lr){3-6} 
 &&Boundary &Number& Graph& \multicolumn{1}{c}{B \& G} &\\

\midrule
 HouseGAN* &  &  & & \checkmark &  & \\
 HouseGAN++ &  &  & &\checkmark &  & \\
 Graph2Plan &  & \checkmark && & \checkmark & \\
 WallPlan & & \checkmark &  &  \checkmark & \checkmark\\
 HouseDiffusion &  &  && \checkmark &  & \\
 Cons2Plan & & \checkmark  & &\checkmark & \checkmark &\\
 GSDiff* & \checkmark & \checkmark& & \checkmark &  & \\
  \rowcolor[gray]{0.9}
 FMLM (Ours) & \checkmark & \checkmark & \checkmark & \checkmark & \checkmark & \checkmark\\
 \bottomrule
\end{tabular}
\end{adjustbox}
\caption{\textbf{Supported tasks.} Our model handles various types of floorplan generation tasks while previous methods perform only specific tasks. * indicates that the models could not generate doors.}
\label{tb:tasks}
\end{table}

\noindent\textbf{Data-driven approaches.}
More recent work focuses on data-driven approaches with generative models~\cite{gan,vae,ddpm}.
% ----------- RASTER-BASED APPROACHES ------------
HouseGAN~\cite{housegan} and HouseGAN++~\cite{houseganpp} adopt convolutional message passing neural networks~\cite{cmpn} to generate raster floorplans from graph conditions.
% ----------- VECTOR-BASED APPROACHES ------------
Graph2Plan~\cite{graph2plan} proposed a coarse-to-fine pipeline that first predicts coarse bounding boxes of rooms and then refines them by simultaneously generated floorplan images.
FloorplanGAN~\cite{floorplangan} introduced a differentiable renderer that rasterizes floorplans images from their vector representations to apply a raster-based GAN framework on vector floorplans.
HouseDiffusion~\cite{housediffusion} proposed a diffusion-based approach that diffuses the positions of room vertices conditioned on room adjacency graphs.
Cons2Plan~\cite{cons2plan} introduces a two-stage model in which the first stage produces graphs from site boundaries, and the subsequent stage generates room vertices in a similar fashion to HouseDiffusion.
GSDiff~\cite{gsdiff} supports variable-length room polygons by introducing a multi-stage pipeline that includes room vertex generation, edge prediction, and room type assignment.

\begin{figure*}[t]
  \centering
  \begin{adjustbox}{width=1.0\linewidth}
  \includegraphics{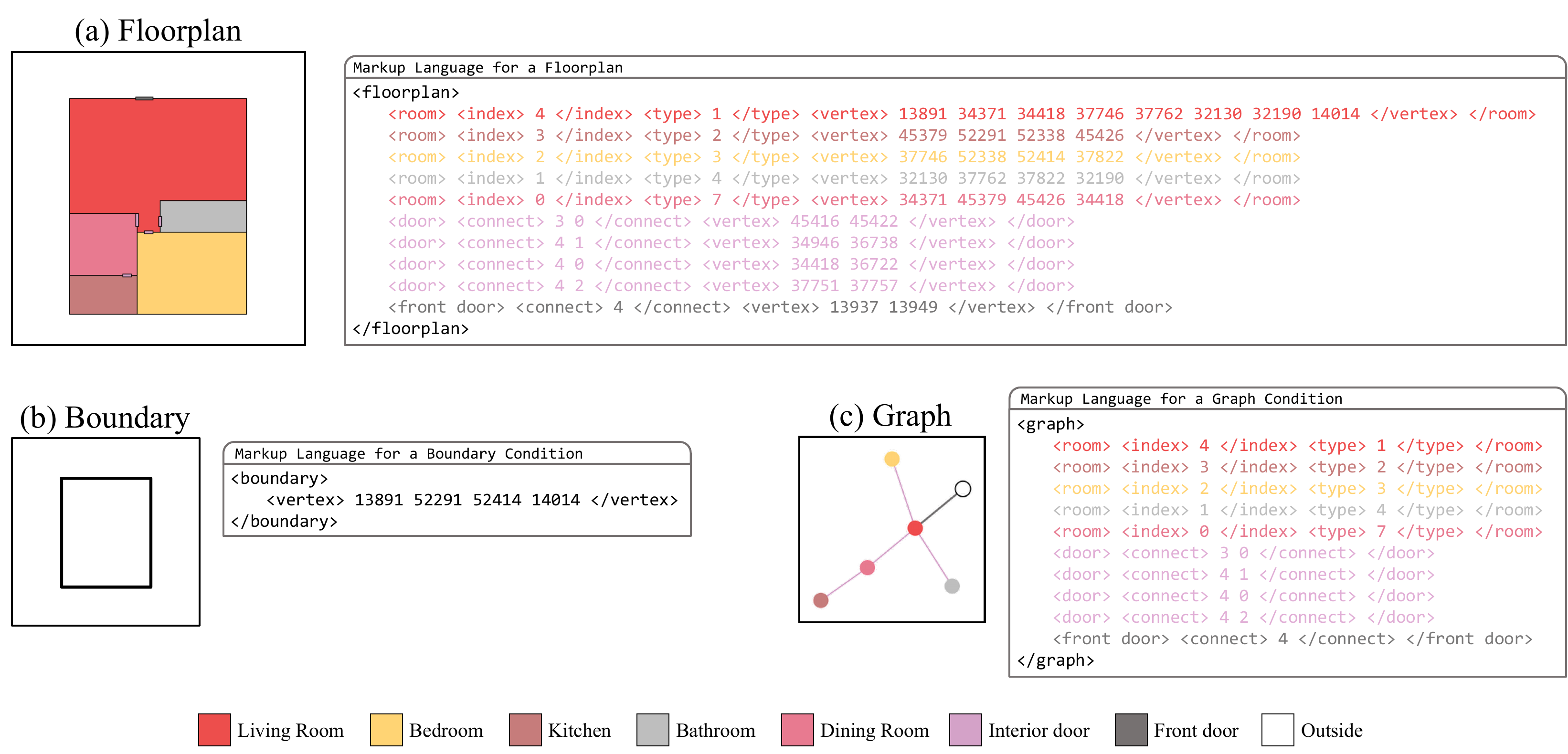}
  \end{adjustbox}
  \caption{\textbf{\repname (\repnameshort).} We represent floorplans, boundaries, and graphs in a markup manner, which unifies the various floorplan generation tasks into a single task of \repnameshort sequence generation.}
  \label{fig:fml}
\end{figure*}

However, these methods suffer from poor generalization ability across different generation tasks due to their suboptimal representations and network designs. 
For example, the recent diffusion-based methods~\cite{housediffusion,gsdiff,cons2plan} mainly focus on the diffusion of the coordinates of room vertices, which forces the system either to pre-condition the graph structure in advance specifying the number of rooms and the number of vertices each room should have~\cite{housediffusion,cons2plan}, or to require additional networks to infer the underlying structures including edge extraction and room type assignment from the generated potential vertices~\cite{gsdiff}. 
Such task-specific network designs make it difficult for the models to generalize to multiple generation tasks as shown in Table~\ref{tb:tasks}.
In contrast, we represent floorplans and conditions such as boundaries and graphs in single sequences, enabling our model to operate in a single stage and generalize to a wide range of floorplan generation tasks.

\subsection{Autoregressive Modeling}
Autoregressive (AR) models generate data sequentially by conditioning each output element on previously generated ones.
Early approaches based on recurrent neural networks (RNNs) such as long short-term memory (LSTM)~\cite{lstm} modeled sequential data such as handwriting synthesis~\cite{graves2013generating} and machine translation~\cite{sutskever2014sequence}.
The Transformer~\cite{attention} further established AR modeling as a general-purpose paradigm for high-dimensional generative modeling, leading to powerful large language models~\cite{attention,gpt,llama3}.
In the computer vision and graphics, autoregressive frameworks have been successfully applied to pixel-level~\cite{pixelcnn}, patch-level~\cite{taming}, and scale-level~\cite{var} image generation, video generation~\cite{magi}, and geometry reconstruction~\cite{cut3r}.

In this work, we base our model on autoregression so that it can generate variable numbers of rooms, doors, and their vertices, taking advantages against non-AR models that generate only fixed numbers of them~\cite{housediffusion,cons2plan} or require additional steps for room allocation~\cite{gsdiff}.

\newcounter{rule} % 
\newcommand{\Rule}{%
  \stepcounter{rule}%
  \underline{Rule \therule}: %
}

\section{Proposed Method}
Our goal is to generate diverse and plausible vector floorplans under various situations such as unconditional generation, conditional generation on the site boundary, the number of rooms, and the room adjacency graph, and completion of partial floorplans.
To achieve this, we introduce a new representation called \repname (\repnameshort) that represents the entire structure information of a floorplan as a single sentence written in a markup language (Sec.~\ref{sec:fml}). 
With \repnameshort, any floorplan generation tasks can be formulated as a sequential generation task of \repnameshort.
Therefore, our introduced simple autoregressive transformer trained to generate \repnameshort sequences, which we named \modelname (\modelnameshort), performs unified vector floorplan generation without raster-based representation~\cite{housegan}, discretization from continuous representation~\cite{housediffusion}, and multi-staging~\cite{gsdiff} (Sec.~\ref{sec:model}).

\subsection{\repname}
\label{sec:fml}
We present \repname (\repnameshort) that represents floorplans and conditions such as boundary and graph in a tag-structured manner as shown in Fig.~\ref{fig:fml}.

\noindent\textbf{Preliminary.}
A floorplan with $N_{r}$ rooms and $N_{d}$ interior doors is represented in 2D space as a set of room polygons with associated room type labels $\{(P_{i},t_i)\}_{i=1}^{N_{r}}$, interior door lines $\{D_{j}\}_{j=1}^{N_d}$, and a front door line $F$ in 2D space.
Each room polygon $P_{i} \in \mathbb{R}^{n_i\times2}$ has a room-specific number of vertices $n_i$ and each interior or front door is represented as a line connecting an initial point and an ending point, \ie, $D_j, F \in \mathbb{R}^{2\times2}$.
The interior doors are placed on the edges between rooms, while the front door is placed on the edge between a room and the outside region.
A boundary condition is represented as a polygon $B \in \mathbb{R}^{n_b\times2}$ with the number of vertices $n_b$. 
A graph condition is represented as an adjacency matrix $G \in \{0,1\}^{N_r \times N_r}$ where each element $G_{i,j}=1$ indicates that rooms $i$ and $j$ are connected via an interior door, and $G_{i,j}=0$ otherwise.

\noindent\textbf{Grammar.}
In \repnameshort, we define four types of tokens including tag, coordinate, room index, and room type. 
We define the grammar of FML as follows:

\noindent\Rule 
FML starts with \verb|<sequence>| and ends with \verb|</sequence>|.

\noindent\Rule 
Between \verb|<sequence>| and \verb|</sequence>|, the tag \verb|<floorplan>| is used to start to describe floorplan information and the tag \verb|</floorplan>| is used to end it.

\noindent\Rule Between \verb|<floorplan>| and \verb|</floorplan>|, the tags \verb|<room>|, \verb|<door>|, and \verb|<front door>| are used to start to describe the information about each room, door, and front door, respectively. 
FML always describes them in the order of rooms $\rightarrow$ doors $\rightarrow$ a front door. 
Also, we use \verb|</room>|, \verb|</door>|, and \verb|</front door>| at the end of the respective descriptions.

\noindent\Rule 
Between \verb|<room>| and \verb|</room>|, FML first places \verb|<index>| and \verb|</index>|. 
A single number between them represents the room index in descending order.

\noindent\Rule 
After \verb|</index>|, FML places \verb|<type>| and \verb|</type>|. 
A single number between them represents the room type (\eg, living room, kitchen, etc.).

\noindent\Rule 
After \verb|</type>|, FML places \verb|<vertex>| and \verb|</vertex>|. 
Multiple numbers between them represent the vertex positions of room polygons. 
Each 2D coordinate $(x,y)$ is converted into a 1D value $z = x + y* W$ where $W$ denotes the width of the 2D space.

\noindent\Rule 
Between \verb|<door>| and \verb|</door>|, FML first places \verb|<connect>| and \verb|</connect>|. 
Two numbers between them represent the room indices that the door connects.

\noindent\Rule 
After \verb|</connect>|, FML places \verb|<vertex>| and \verb|</vertex>|. 
Two numbers between them represent the beginning and ending coordinates of the door. 
We use the same coordinate space as room polygons.

\noindent\Rule 
Front doors are described in the same manner as doors by \verb|<front door>| and \verb|</front door>|, except that only a single number is placed between \verb|<connect>| and \verb|</connect>| because a front door connects a single room to the outside.

In addition to the basic rules above to represent floorplans, we define the following rules to support conditional generation tasks such as boundary and graph as follows:

\noindent\Rule 
Before the floorplan description, FML can place tags \verb|<condition>| and \verb|</condition>| that correspond to the beginning and ending of condition information, respectively. 
FML can use \verb|<boundary>| and \verb|<graph>| to start to describe the information about boundary and graph condition, respectively. 
FML always describes in the order of boundary $\rightarrow$ graph. 
Also, we use \verb|</boundary>| and \verb|</graph>| at the end of the respective descriptions.

\noindent\Rule 
Between \verb|<boundary>| and \verb|</boundary>|, FML describes a boundary condition by the set of boundary vertices in the same manner as room vertices. 

\noindent\Rule 
Between \verb|<graph>| and \verb|</graph>|, FML represents a room adjacency graph by the same manner as floorplans but without vertices information.

\begin{figure}[t]
  \centering
  \begin{adjustbox}{width=1.0\linewidth}
  \includegraphics{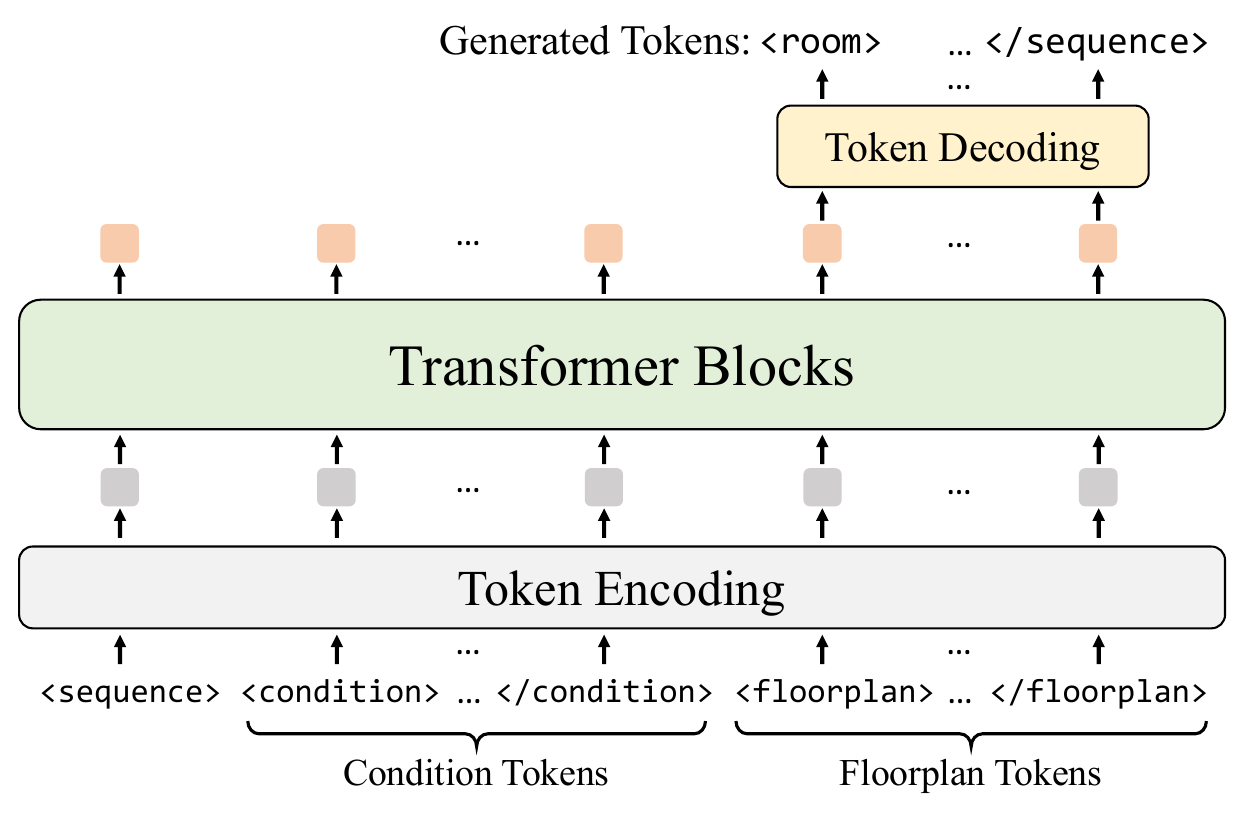}
  \end{adjustbox}
  \caption{\textbf{Overview of \modelnameshort.} Similarly to LLMs, \modelnameshort is based on a simple transformer model that is trained by next token prediction and performs autoregressive inference. }
  \label{fig:architecture}
\end{figure}

\subsection{\modelname}
\label{sec:model}
We introduce \modelname (\modelnameshort) to generate \repnameshort sequences.
The overview of the architecture is illustrated in Fig.~\ref{fig:architecture}.
Our model is based on an autoregressive transformer that takes a token sequence and predicts its next token.
The transformer block is the same architecture as LLaMA-3~\cite{llama3} which consists of self-attention~\cite{attention}, layer normalization~\cite{layernorm}, and multi-layer perceptron.
We encode and decode tokens as follows: 

\noindent\textbf{Encoding.}
We encode tags, room indices, and room types by assigning a learnable vector to each class.
For coordinates, we encode 2D values into a single vector by sinusoidal positional embedding and apply a learnable linear projection to it.

\noindent\textbf{Decoding.}
To decode the markup language from output tokens, we apply a learnable linear projection head $W \in \mathbb{R}^{(C_{\text{tag}} + C_{\text{coord}} + C_{\text{index}} + C_{\text{type}})\times C}$ where $C_{\text{tag}}$ is the number of tag classes, $C_{\text{tag}}=H*W$ with the height $H$ and width $W$ of the 2D coordinate space, $C_{\text{index}}$ is the maximum number of rooms per floorplan in the dataset,  $C_{\text{type}}$ is the number of room types, and $C$ is the dimension size of the output embedding.
Let $\boldsymbol{x}=\{x_i\}_{i=1}^{L}$ be an input \repnameshort sequence with a length of $L$ and $f$ be the transformer that takes $\{x_i\}_{i=1}^{l}$ and outputs embedding $f(\{x_i\}_{i=1}^{l}) \in \mathbb{R}^{C}$ in the teacher-forcing manner to predict the next token $x_{l+1}$.
The embedding $f(\{x_i\}_{i=1}^{l})$ is cast into the class probability with the head $W$ and the softmax operation:
\begin{equation}
p(x_{l+1}\mid \{x_i\}_{i=1}^{l}) = \text{softmax}(Wf(\{x_i\}_{i=1}^{l})).
\end{equation}

\begin{figure}[t]
  \centering
  \begin{adjustbox}{width=1.0\linewidth}
  \includegraphics{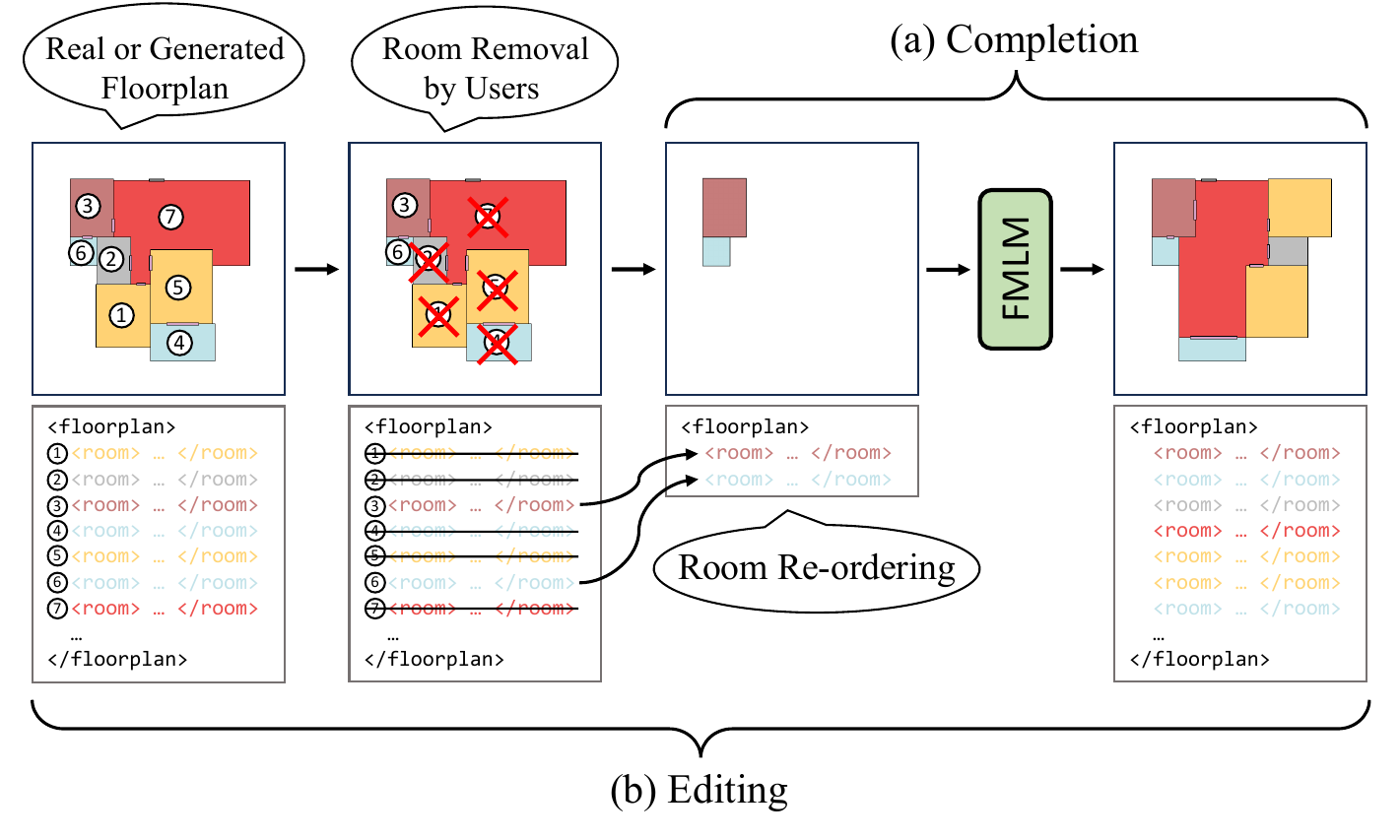}
  \end{adjustbox}
  \caption{\textbf{Floorplan completion and editing.} (a) Our model complements incomplete floorplans by just starting with an incomplete sequence. (b) With this capability, our model can be incorporated into interactive editing with users.}
  \label{fig:editing}
\end{figure}

\subsection{Training}
\label{sec:training}
We train our model with the cross-entropy loss:
\begin{equation}
\label{eq:loss}
\mathcal{L} = \mathop{\mathbb{E}}_{\boldsymbol{x} \sim \mathcal{D}} \left[ - \sum_{l=l_1}^{L-1} \log p(x_{l+1} \mid \{x_i\}_{i=1}^{l})  \right],
\end{equation}
where $\mathcal{D}$ is the dataset and $l_1$ denotes the token index of \verb|<floorplan>| in $\boldsymbol{x}$; we apply the loss only on the floorplan tokens as shown in Fig.~\ref{fig:architecture}.

Unlike recent diffusion-based models~\cite{housediffusion,cons2plan}, our model is originally permutation-sensitive for room orders, which is incompatible with the permutation-equivalent nature of floorplans.
% 
% Therefore, it is important to learn permutation-equivalency for robust floorplan generation.
% 
Therefore, we introduce a simple yet effective data augmentation technique for \modelnameshort, called room permutation, that randomizes the order of rooms in floorplans to learn permutation-equivalency.

\subsection{Constrained Decoding}
To generate consistent floorplans during inference, we force our model to strictly follow the grammar of \repnameshort by setting the probabilities of improper classes to 0. 
For example:
\begin{enumerate}
\item Doors should have just two vertices.
\item Room vertices should be placed outside the previously generated rooms. 
\end{enumerate}
The full list is found in appendix.
Enabling integration of such heuristic rules into the generator is one of the advantages of autoregressive modeling compared to the previous non-AR generative approaches~\cite{houseganpp,housediffusion,gsdiff}.

\subsection{Unconditional Generation}
Once inputting tags \verb|<sequence>| \verb|<floorplan>| into our model, it autoregressively generates subsequent tokens without any conditions.

\subsection{Conditional Generation}

\noindent\textbf{Number conditions.}
Our model naturally performs conditional generation by the number of rooms using the \verb|<index>| tag.
Concretely, when we aim to generate a room that has five rooms, we start autoregressive generation with ``\verb|<sequence>| \verb|<floorplan>| \verb|<room>| \verb|<index>| 4''.

\noindent\textbf{Boundary conditions.}
As shown in Fig.~\ref{fig:fml}(b), we input a sequence of vertex coordinates of the boundary.

\noindent\textbf{Graph conditions.}
As shown in Fig.~\ref{fig:fml}(c), we represent a graph condition that constrains the number and types of rooms.
To encourage our model to learn the correlation between graph conditions and target floorplans, we sort the rooms in each ground truth floorplan to match the order in the corresponding graph condition.

\subsection{Completion and Editing.}
Our model is also capable of floorplan completion and editing as follows:

\noindent\textbf{Completion.}
As shown in Fig.~\ref{fig:editing}(a), by starting with an incomplete floorplan sequence, our model complements the sequence to generate its full floorplan.

\noindent\textbf{Editing.}
Floorplan editing is performed by combining a removal process with completion as shown in Fig.~\ref{fig:editing}(b).

\begin{table}[t]
\centering
\begin{adjustbox}{width=0.7\linewidth}
\begin{tabular}{c|c|cc}
\toprule
Condition & Method & FID ($\downarrow$) & IoU ($\uparrow$)\\
\midrule
\multirow{2}{*}{Boundary}
 & Graph2Plan & 34.20 &95.87\\
 & \modelnameshort (Ours) & \textbf{6.51} & \textbf{97.86}\\
 \bottomrule
\end{tabular}
\end{adjustbox}
\caption{\textbf{Comparison with Graph2Plan on boundary condition.}}
\label{tb:g2p_boundary}
\end{table}

\begin{table}[t]
\centering
\begin{adjustbox}{width=0.7\linewidth}
\begin{tabular}{c|c|cc}
\toprule
Condition & Method & FID ($\downarrow$) & IoU ($\uparrow$)\\
\midrule
\multirow{2}{*}{Boundary}
 & GSDiff &11.26 & 97.21 \\
 & \modelnameshort (Ours) & \textbf{4.61} & \textbf{98.06}\\
 \bottomrule
\end{tabular}
\end{adjustbox}
\caption{\textbf{Comparison with GSDiff on boundary condition.}}
\label{tb:gsdiff_boundary}
\end{table}

\begin{figure*}[t]
  \centering
  \begin{minipage}{1.0\linewidth}
  \centering
  \includegraphics[width=1.0\linewidth]{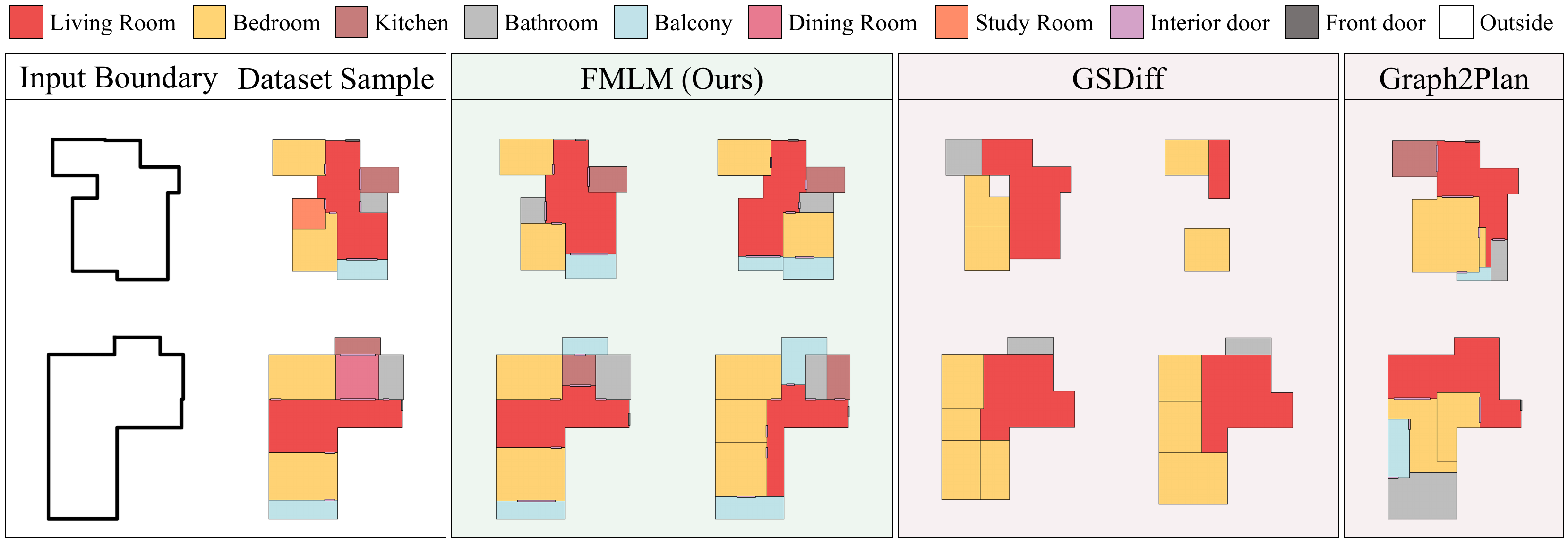}
  \subcaption{Boundary-conditional generation.}
  \label{fig:qualitative_boundary}
  \end{minipage}
  \par\vspace{1em}
  \begin{minipage}{1.0\linewidth}
  \centering
  \includegraphics[width=1.0\linewidth]{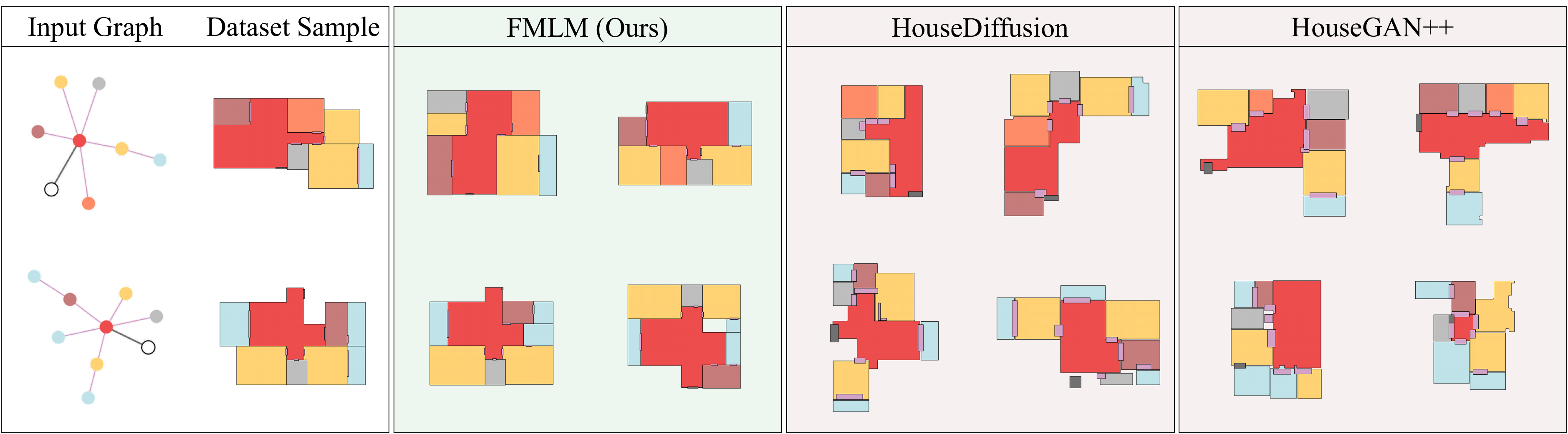}
  \subcaption{Graph-conditional generation}
  \label{fig:qualitative_graph}
  \end{minipage}
  \par\vspace{1em}
  \begin{minipage}{1.0\linewidth}
  \centering
  \includegraphics[width=1.0\linewidth]{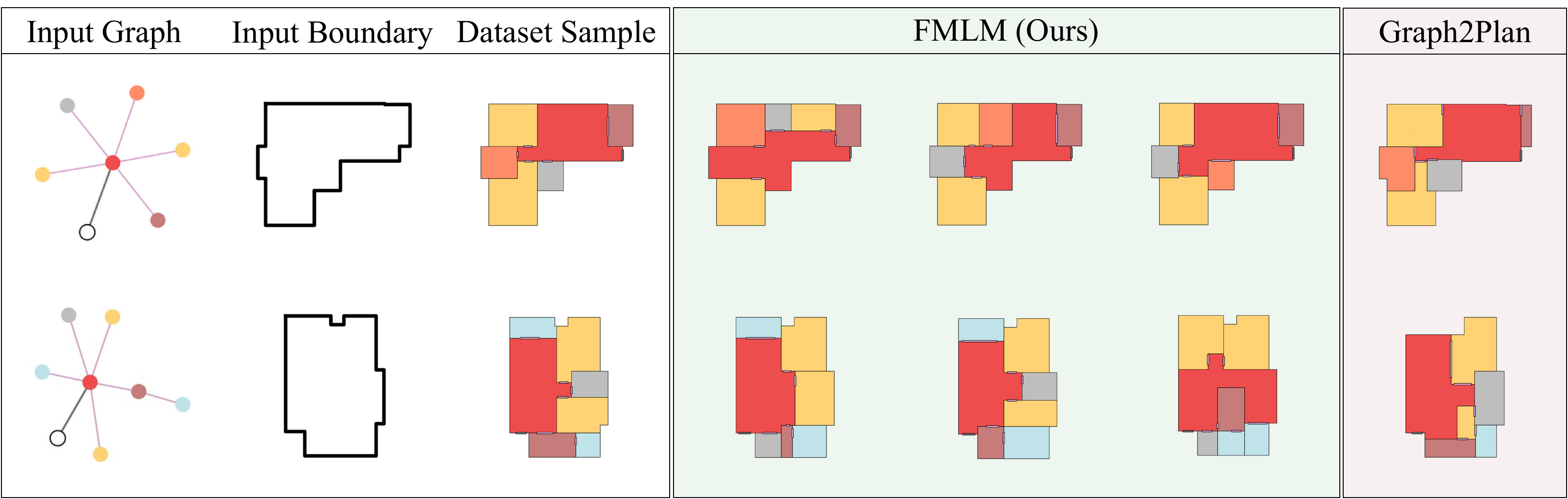}
  \subcaption{Multi-conditional generation}
  \label{fig:qualitative_boundary_graph}
  \end{minipage}
  \caption{\textbf{Qualitative comparison with Graph2Plan and HouseDiffusion.} Our model generates more consistent and realistic examples from various types of conditions, (a) boundary, (b) graph, (c) boundary and graph, than task-specific models such as Graph2Plan and HouseDiffusion. Note that Graph2Plan generates only a single floorplan for each condition. Best viewed in zoom.} 
  \label{fig:qualitative}
\end{figure*}

\begin{table*}[t]
\centering
\begin{adjustbox}{width=0.75\linewidth}
\begin{tabular}{c|c|cccc|ccccc|c}
\toprule
\multirow{2}{*}{Condition} & \multirow{2}{*}{Method} & 
\multicolumn{5}{c}{FID ($\downarrow$)} & \multicolumn{5}{c}{GED ($\downarrow$)}\\
\cmidrule(lr){3-7} \cmidrule(lr){8-12}
&&5&6&7&8&ALL&5&6&7&8&ALL\\
\midrule
\multirow{3}{*}{Graph}
 & HouseGAN++ & 
44.63 & 46.88 & 49.62 & 54.49 & 48.44 &
 1.56 & 2.04 & 2.51 & 3.25 & 2.57\\
 & HouseDiffusion & 
32.98 & 27.15 & 31.61 & 30.43 & 29.31 &
 0.97 & 1.34 & 1.52 & \textbf{1.87} & 1.55\\
 & \modelnameshort (Ours) & 
\textbf{6.97} & \textbf{5.20} & \textbf{4.07} & \textbf{4.64} & \textbf{3.41} &
 \textbf{0.49} & \textbf{0.69} & \textbf{1.13} & 1.96 & \textbf{1.21}\\
\bottomrule
\end{tabular}
\end{adjustbox}
\caption{\textbf{Comparison with HouseDiffusion on graph condition.}}
\label{tb:housediff_graph}
\end{table*}

\begin{table*}[t]
\centering
\begin{adjustbox}{width=1.0\linewidth}
\begin{tabular}{c|c|cccc|ccccc|ccccc|c}
\toprule
\multirow{2}{*}{Condition} & \multirow{2}{*}{Method} & \multicolumn{5}{c}{FID ($\downarrow$)} & \multicolumn{5}{c}{GED ($\downarrow$)} & \multicolumn{5}{c}{IoU ($\uparrow$)}\\
\cmidrule(lr){3-7} \cmidrule(lr){8-12} \cmidrule(lr){13-17}
&&5&6&7&8&ALL&5&6&7&8&ALL&5&6&7&8&ALL\\
\midrule
\multirow{2}{*}{B \& G}
 & Graph2Plan  & 
 44.32 & 28.84 & 29.53 & 36.07 &  22.87 &
 2.30 & 2.95 & 3.55 & 4.26 & 3.43 & 
 93.72 & 93.56 & 92.69 & 92.29 & 92.96\\
 & \modelnameshort (Ours)  &  
\textbf{29.20} & \textbf{14.88} & \textbf{21.85} & \textbf{30.64} & \textbf{14.17} &
 \textbf{0.45} & \textbf{0.71} & \textbf{1.30} & \textbf{2.01} & \textbf{1.24} &
 \textbf{98.42} & \textbf{98.31} & \textbf{97.52} & \textbf{96.64} & \textbf{97.59}\\
\bottomrule
\end{tabular}
\end{adjustbox}
\caption{\textbf{Comparison with Graph2Plan on boundary-graph condition.}}
\label{tb:g2p_boundary_graph}
\end{table*}

\section{Experiments}

\subsection{Setup}
\noindent\textbf{Dataset.}
We used RPLAN dataset~\cite{rplan} that comprises 80k floorplans with dense annotations. 
Following previous work~\cite{houseganpp,housediffusion}, our model is trained to generate nine room classes \ie, \textit{Living Room}, \textit{Bedroom}, \textit{Bathroom}, \textit{Dining Room}, \textit{Kitchen}, \textit{Study Room}, \textit{Entrance}, \textit{Storage}, and \textit{Balcony}, and two door classes, \ie, \textit{Interior Door} and \textit{Front Door}.

\noindent\textbf{Metrics.}
Commonly on all the floorplan generation tasks, we use Frechet Inception
Distance (FID)~\cite{fid} to evaluate the distribution distance between real floorplan images and generated ones.
On boundary conditional generation tasks, we introduce Intersection over Union (IoU) to evaluate the distance between internal regions of boundaries and union of generated room regions.
On graph conditional generation tasks, we adopt Graph Edit Distance (GED)~\cite{ged} to evaluate the distance between condition graphs and reconstructed ones from generated floorplans. 

\noindent\textbf{Baselines.}
We compare our model with the previous approaches including HouseGAN++~\cite{houseganpp}, HouseDiffusion~\cite{housediffusion}, Graph2Plan~\cite{graph2plan}, and GSDiff~\cite{gsdiff}.

\noindent\textbf{Implementation details.}
We implement our model with PyTorch~\cite{pytorch}.
All the experiments are conducted on a single NVIDIA A100 GPU.
$C_{\text{tag}}$, $C_{\text{index}}$, $C_{\text{type}}$, $H$, and $W$ are set to 15, 8, 9, 256, and 256, respectively.
Note that we exclude \verb|<sequence>|, \verb|<condition>|, \verb|</condition>|, \verb|<boundary>|, \verb|</boundary>|, \verb|<graph>|, \verb|</graph>|, and \verb|<floorplan>| from the tag classes predicted by the model because they never appear as targets in Eq.~\ref{eq:loss}.
We train our model from scratch for 50 epochs with Adam~\cite{adam} optimizer.
The batch size and learning rate are set to 32 and $1.0^{-4}$, respectively.
The boundary and graph conditions are dropped out by a chance of 50\%, respectively.
Therefore, our model is trained without conditions, with only the boundary condition, with only the graph condition, and with both boundary and graph conditions, each with a probability of 25\%.
More details are included in appendix.

\subsection{Unconditional Generation}
\label{sec:exp_uncond}
We compare our model with GSDiff~\cite{gsdiff} on unconditional generation.
Since GSDiff could not generate doors, we remove doors in our predicted floorplans and ground truth ones during computing FID for a fair comparison.
We observe that our method achieves \textbf{7.22} while GSDiff achieves 15.02, indicating that our method outperforms GSDiff.

\subsection{Conditional Generation}
\noindent\textbf{Boundary.}
We evaluate our model on the boundary conditional generation task in comparison to Graph2Plan~\cite{graph2plan} and GSDiff~\cite{gsdiff}.
We give the numerical comparisons with Graph2Plan in Table~\ref{tb:g2p_boundary} and with GSDiff in Table~\ref{tb:gsdiff_boundary}.
Note that because the official pre-trained models of Graph2Plan and GSDiff adopted different train/test splits, we train our model on each training set and evaluate them on each test set for fair comparison.
We show the generated examples in Fig.~\ref{fig:qualitative_boundary}.
We can see that our model is more consistent with the boundary conditions than Graph2Plan which often lacks interior doors to connect rooms.
Also, our method can generate more plausible floorplans than GSDiff that lacks layout fidelity and  diversity.
Our method outperforms them in both FID and IoU, indicating our model generates higher-fidelity floorplans from boundary conditions.

\noindent\textbf{Graph.}
We then evaluate our model on the graph conditional generation task in comparison to HouseGAN++~\cite{houseganpp} and HouseDiffusion~\cite{housediffusion}.
To prevent models from simply outputting the same examples as in training data, we first divided floorplans into five groups depending on the number of rooms, \ie, five, six, seven, and eight rooms. 
Then, for each group, we divided the samples into training and testing sets so that identical graphs do not overlap between training and testing. 
We train our model, HouseGAN++~\cite{houseganpp}, and HouseDiffusion~\cite{housediffusion} on the combined training samples of all groups.
Note that we could not compare GSDiff~\cite{gsdiff} as it does not generate doors and, therefore, the definition of room adjacency of GSDiff is different from that of our model, HouseGAN++, and HouseDiffusion.
We show the generated examples in Fig.~\ref{fig:qualitative_graph}.
We can see that all the methods succeed in generating plausible floorplans from graph conditions. 
However, HouseGAN++ and HouseDiffusion sometimes place interior doors in locations that are not the boundaries between rooms, which does not occur in our method because of constrained decoding.
We also give the quantitative comparison in Table~\ref{tb:housediff_graph}.
Similarly with Table~\ref{tb:gsdiff_boundary}, we remove doors when we compute FID for a fair comparison because there is a difference in door representations, where HouseGAN++ and HouseDiffusion represent doors by polygons while our method represents them by lines.
Our model outperforms HouseGAN++ and HouseDiffusion in all the metrics except GED on eight rooms where our model slightly underperforms HouseDiffusion. 
This may be because the number of samples with eight rooms in the training set is much fewer than the others, which can be improved by some techniques such as weighted sampling during training. 

\noindent\textbf{Boundary and Graph.}
We here evaluate our model conditioned by both boundary and graph.
We used the same checkpoints of our model and Graph2Plan as in boundary-conditional generation.
Because Graph2Plan requires the front door positions in advance while our method does not, we exclude the effect of front doors when we evaluate GED.
The generated examples of our model and Graph2Plan can be found in Fig.~\ref{fig:qualitative_boundary_graph}.
Our model generates high-fidelity floorplans from graph and boundary conditions while Graph2Plan tends to generate unnatural layouts and fails to generate expected room types.
We also give the quantitative comparison in Table~\ref{tb:g2p_boundary_graph}, where our approach outperforms Graph2Plan in all the metrics including FID, GED, and IoU for all the numbers of rooms.

\noindent\textbf{Number.}
Due to space limitations, we show the generated examples conditioned by the number of rooms in appendix.

\begin{figure}[t]
  \centering
  \begin{adjustbox}{width=1.0\linewidth}
  \includegraphics{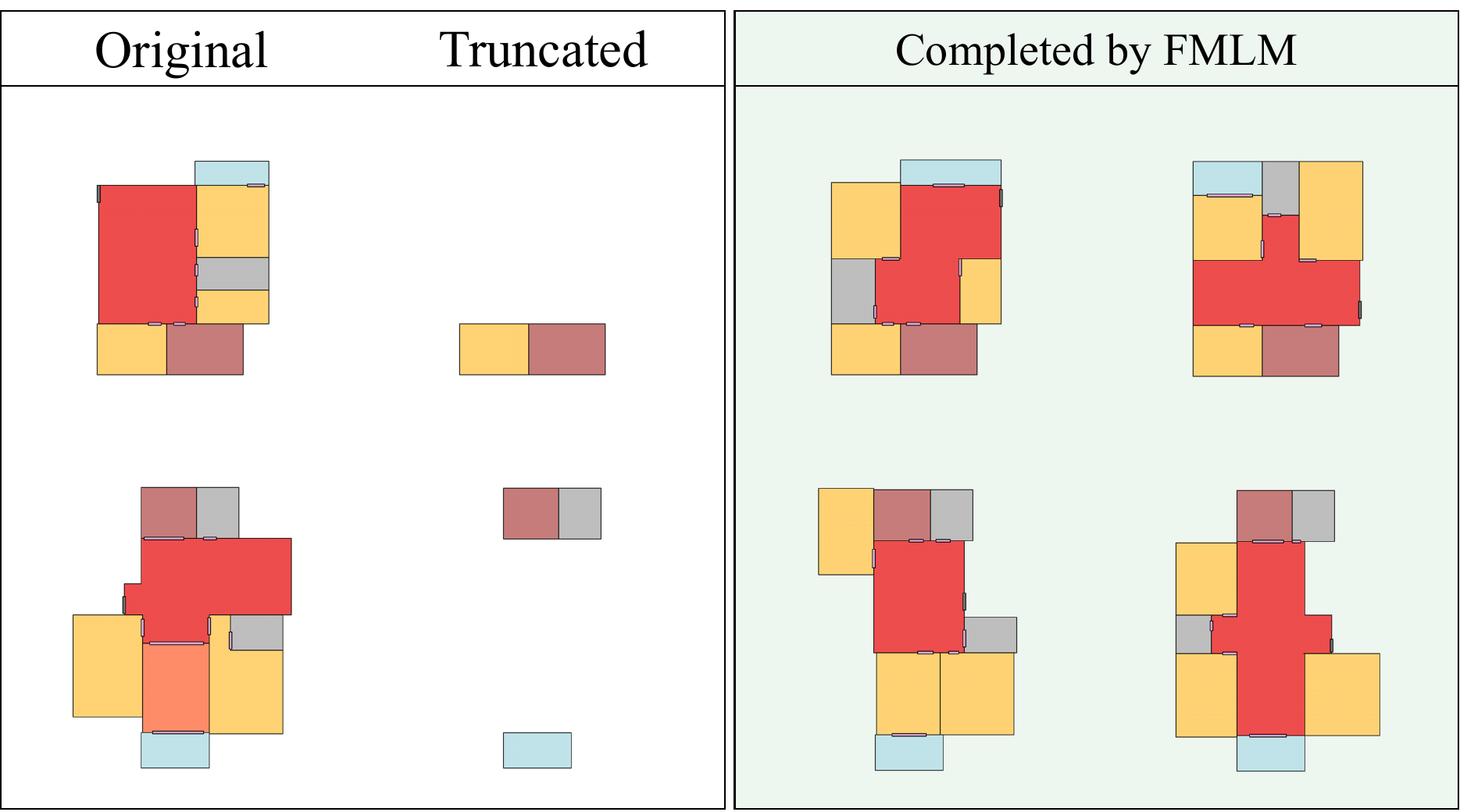}
  \end{adjustbox}
  \caption{\textbf{Floorplan completion.}}
  \label{fig:qualitative_completion}
\end{figure}

\begin{table}[t]
\centering
\begin{adjustbox}{width=0.9\linewidth}
\begin{tabular}{c|c|ccc}
\toprule
Condition & Setting & FID ($\downarrow$) & GED ($\downarrow$)& IoU ($\uparrow$)\\
\midrule
\multirow{2}{*}{B \& G}
 & w/o Permutation  & 24.36 & 2.35 &95.82 \\
 & Ours  & \textbf{14.17} & \textbf{1.24} & \textbf{97.59}\\
\bottomrule
\end{tabular}
\end{adjustbox}
\caption{\textbf{Effect of room permutation.}}
\label{tb:ablation_perm}
\end{table}

\begin{table}[t]
\centering
\begin{adjustbox}{width=1.0\linewidth}
\begin{tabular}{c|c|cccc|c}
\toprule
\multirow{2}{*}{Condition} & \multirow{2}{*}{Setting} &\multicolumn{5}{c}{FID ($\downarrow$)}\\
\cmidrule(lr){3-7} 
&&5&6&7&8&ALL\\
\midrule
\multirow{2}{*}{Number}
 & Ascending Order  & 105.78 & 112.29 &  114.80 & 125.13 &  94.57  \\
 & Ours  &  \textbf{47.17} & \textbf{46.89} & \textbf{49.38} & \textbf{48.82} & \textbf{25.50}\\
\bottomrule
\end{tabular}
\end{adjustbox}
\caption{\textbf{Effect of indexing in descending order.}}
\label{tb:ablation_index}
\end{table}

\subsection{Completion}
Our model is also capable of the completion of partial floorplans.
As shown in Fig.~\ref{fig:qualitative_completion}, our model generates diverse samples from the same part of rooms.
This result indicates that our method empowers users to interactively edit floorplans.

\subsection{Ablation Studies}
\noindent\textbf{Effect of room permutation.}
We compare our full model and its variant without room permutation in Table~\ref{tb:ablation_perm}.
The augmentation improves all the metrics, demonstrating the importance of learning permutation-equivalency.

\noindent\textbf{Effect of indexing in descending order.}
Another key design of \repnameshort is to describe room indices in descending order.
To demonstrate this, we train a variant where room indices are described in ascending order and compare it with our model on number-conditional generation in Table~\ref{tb:ablation_index}.
For the variant, we stop generating rooms once the number of the generated rooms reaches the conditioned number.
The variant significantly drops FID; this is because, in ascending order setting, the model could not determine in advance how many rooms should be generated.

\begin{figure}[t]
  \centering
  \begin{adjustbox}{width=1.0\linewidth}
  \includegraphics{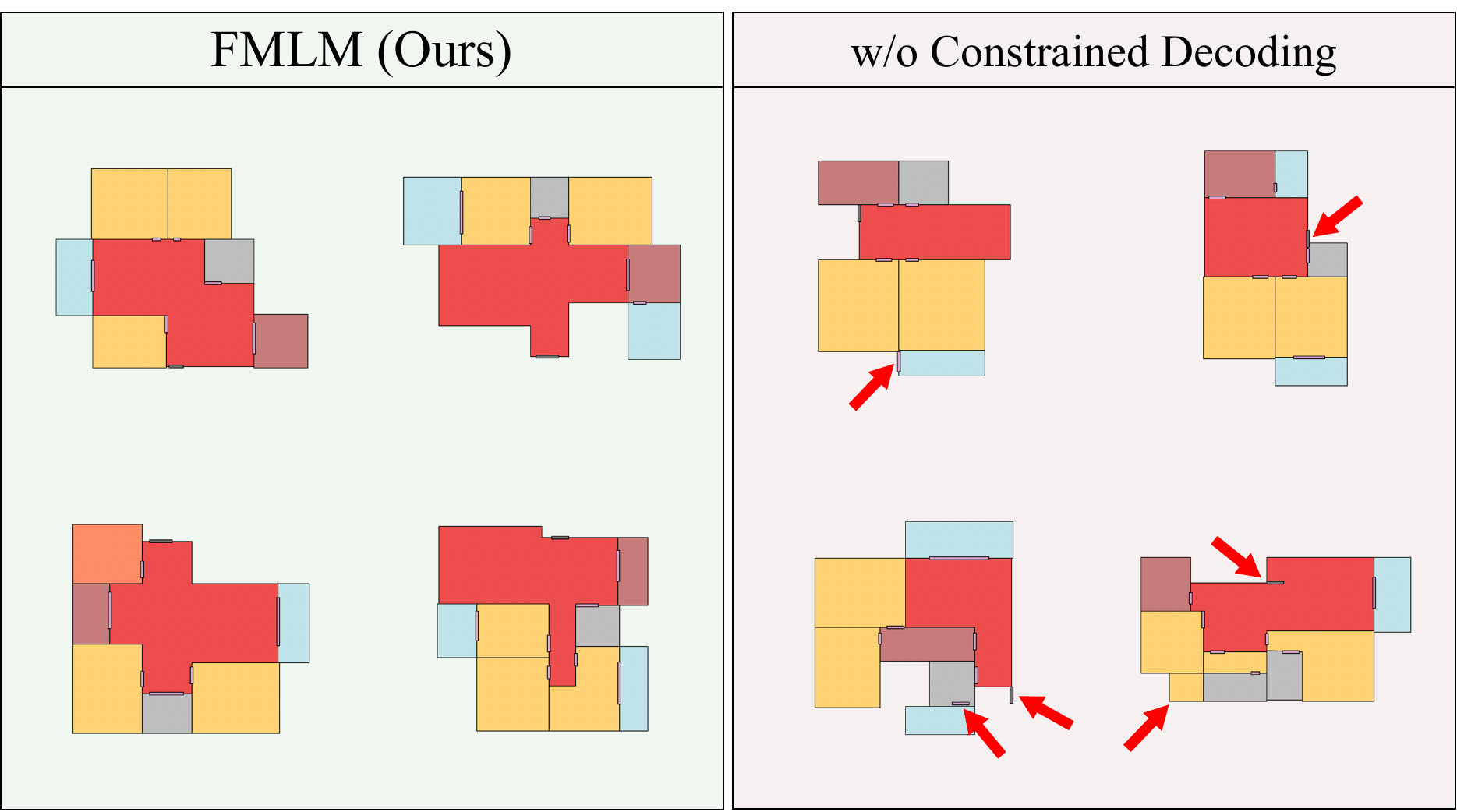}
  \end{adjustbox}
  \caption{\textbf{Effect of constrained decoding.} Best viewed in zoom.}
  \label{fig:ablation_constrain}
\end{figure}

\noindent\textbf{Effect of constrained decoding.}
We also show the effect of constrained decoding in Fig.~\ref{fig:ablation_constrain} by comparing the cases with and without it during inference.
We generate rooms without any conditions by two models where one of them is our full model and the other is our model without constrained decoding.
We can see that without it, floorplans are often generated with overlapping rooms and misaligned doors that appear outside the edges between rooms.

\section{Limitations and Future Work}
Our approach still has some limitations that point to promising directions for future research.
First, the integration of \repnameshort into LLMs presents an opportunity to enable floorplan generation directly from natural-language descriptions.
This direction could substantially broaden the accessibility and controllability of layout synthesis. 
Second, our current model is restricted to single-story floorplans. 
Extending \modelnameshort to handle multi-story designs is a natural next step, which may be achieved by introducing additional structural tags such as \verb|<story>| and \verb|</story>| to represent vertical hierarchy and inter-floor relationships.

\section{Conclusion}
In this paper, we present \repname (\repnameshort) for unified vector floorplan generation.
\repnameshort describes floorplans and constraints such as site boundaries and room adjacency graphs in a markup manner, which unifies various floorplan generation tasks into a \repnameshort sequence generation task.
Extensive experiments show that our autoregressive transformer model trained with \repnameshort outperforms the previous methods designed for specific tasks, demonstrating the strong generalization ability of our representation across different generation tasks.
\clearpage
\section*{Acknowledgments}
This work was partially financially supported by JST ASPIRE Program, Japan, Grant Number JPMJAP2303.

{
    \small
    \bibliographystyle{ieeenat_fullname}
    \bibliography{main}
}

\clearpage
\maketitlesupplementary
\appendix

\section{More Implementation Details}
\noindent\textbf{Hyper-parameters.}
We give the additional information about the hyper-parameters of our model in Table~\ref{tb:hyperparam}.
\begin{wraptable}{r}{0.45\columnwidth}
    \centering
    \vspace{-2mm}
    \begin{adjustbox}{width=1.0\linewidth}
    \begin{tabular}{c|c}
        \toprule
        Hyperparameter & Value\\
        \midrule
        Transformer Dimension & 512\\
        MLP Dimension &2048 \\
        Num Heads & 32 \\
        Num Layers & 24 \\
        Temperature & 0.6 \\ 
        Top P & 0.8 \\
        \bottomrule
    \end{tabular}
    \end{adjustbox}
    \caption{\textbf{Hyper-parameters.}}
    \label{tb:hyperparam}
    \vspace{-4mm}
\end{wraptable}
\textit{Temperature} and \textit{Top P} are the hyper-parameters used to determine the predicted class from the probability distribution with randomness, seen frequently in the context of LLMs.
We introduce such randomness when the variation is required, as shown in Table~\ref{tb:random}.

\begin{table}[h]
\centering
\begin{adjustbox}{width=1.0\linewidth}
\begin{tabular}{c|cccccc}
\toprule
Token Type &  Uncond. & Boundary& Number & Graph & G\&B & Compl.\\
\midrule
Tag & & & \\
Coordinate & \checkmark & \checkmark & \checkmark & \checkmark & \checkmark & \checkmark\\
Room index & \checkmark & \checkmark &  & & & \checkmark \\
Room type & \checkmark & \checkmark &\checkmark & & &\checkmark \\
\bottomrule
\end{tabular}
\end{adjustbox}
\caption{\textbf{Randomness.}}
\label{tb:random}
\vspace{-2mm}
\end{table}

To give the randomness to specific token types, we determine which token type should appear next by the following process:
1) If the token type can be uniquely determined by the grammar of \repnameshort, we adopt it.
2) If not, we compute the sum of probabilities of each token type and we adopt the token type whose sum is the largest.
Once the token type determined, we sample the class only from the type.

\noindent\textbf{Dataset size.}
In Table~\ref{tb:size}, we give the numbers of train/test samples in our experiments in the main paper.
\begin{table}[h]
\centering
\begin{adjustbox}{width=1.0\linewidth}
\begin{tabular}{ccccccc}
\toprule
 Sec.~\ref{sec:exp_uncond}  & Table~\ref{tb:g2p_boundary} & Table~\ref{tb:gsdiff_boundary} & Table~\ref{tb:housediff_graph} & Table~\ref{tb:g2p_boundary_graph} & Table~\ref{tb:ablation_perm} & Table~\ref{tb:ablation_index} \\
\midrule
 65763 / 100 & 74995 / 2880 & 65763 / 378 & 59232 / 9895 & 74995 / 2880 & 74995 / 2880 & 74995 / 400\\
\bottomrule
\end{tabular}
\end{adjustbox}
\caption{\textbf{Dataset size.}}
\label{tb:size}
\end{table}

\noindent\textbf{Constrains.}
We list the constrains in decoding in Table~\ref{tb:constrains}.

\begin{table}[h]
    \centering
    \begin{adjustbox}{width=1.0\linewidth}
    \begin{tabular}{p{8cm}}
    \toprule
    \begin{minipage}[t]{\linewidth}
    \begin{enumerate}[
        label=\arabic*.,
        leftmargin=*,
        topsep=0pt,    
        partopsep=0pt,
        parsep=0pt,
        itemsep=0.3ex  
    ]
      \vspace{-2.5mm}
      \item An interior and front door should have just two vertices.
      \item Room vertices should be placed outside the previously generated rooms.
      \item Interior door vertices should be placed on a edge between two different rooms.
      \item Front door vertices should be placed on a edge between a room and the outside region.
      \item A room should have four or more vertices.
    \end{enumerate}
    \end{minipage}\\
    \addlinespace[8pt]
      \bottomrule
    \end{tabular}
\end{adjustbox}
    \caption{\textbf{Constrains used in decoding.}}
    \label{tb:constrains}
\end{table}

\noindent\textbf{Pre-processing.}
We follow the pre-processing code provided by HouseGAN++.
Also, since FML requires that every two rooms supposed to be adjacent must share an edge for a door, we inflated the room polygons so that they have the shared edge.
We also re-computed the adjacency graph to filter out incorrect annotations after inflation.

\section{User Study}
For further evaluation, we conduct a user study on Amazon Mechanical Turk.
We show 20 users the generated floorplans and ask them to ``select the most functional and natural floorplan''.
We uniformly pick 100 sets of generated results from our model, HouseGAN++, and HouseDiffusion on the graph conditional generation task.
Table~\ref{tb:userstudy} shows the winning rate, \ie, the percentage of cases each method was selected as the most functional and natural floorplan by the users. 
Note that ties can occur when multiple methods receive the same number of votes, so the total does not necessarily sum to 100\%.
We can see that our method is much more preferable than the previous methods, indicating our superior functionality and naturalness.

\begin{table}[h]
% \vspace{-2mm}
\centering
\begin{adjustbox}{width=0.7\linewidth}
\begin{tabular}{ccc}
\toprule
 Ours  & HouseGAN++ & HouseDiffusion\\
\midrule
\textbf{51\%} (51/100) & 24\% (24/100) & 32\% (32/100)\\
\bottomrule
\end{tabular}
\end{adjustbox}
\caption{\textbf{Winning rate on the user study.}}
\label{tb:userstudy}
\end{table}

\section{More Experiments}
\noindent\textbf{Effect of multi-task learning.}
In Table~\ref{tb:multitask}, we train additional variants by dropping out some learning tasks and evaluate them on the graph-conditional task.
The result gives us an interesting finding: the task-specific variant (a) achieves the best GED, and the variant trained on uncond.\&graph (b) significantly worsens GED; however, the more tasks we add, the better GED we obtain, as observed in (c) and (d).

\begin{table}[h]
    \small
    \centering
    \vspace{-2mm}
    \begin{adjustbox}{width=0.95\linewidth}
        \begin{tabular}{c|ccccc|c} \toprule
        \multirow{2}{*}{Evaluation Task} & \multirow{2}{*}{Setting} &  \multicolumn{4}{c}{Learning Tasks} & \multirow{2}{*}{GED ($\downarrow$)} \\ 
        \cmidrule(lr){3-6}
          &&Uncond. & Boundary & Graph &B \& G & \\
        \midrule
         \multirow{4}{*}{Graph}& (a) & &  & \checkmark &  &  0.99 \\
        & (b)&\checkmark &  & \checkmark &  & 1.41 \\
        & (c)&\checkmark & \checkmark & \checkmark &  & 1.34\\
        & (d)&\checkmark & \checkmark & \checkmark & \checkmark & 1.21  \\
        \bottomrule
\end{tabular}
    \end{adjustbox}
  \caption{\textbf{Effect of multi-task learning.}}
  
  \label{tb:multitask}
\end{table}

\noindent\textbf{Computation cost.}
In Table~\ref{tb:complexity}, we compare our model with HouseGAN++ and HouseDiffusion in terms of training time and per-sample inference time for each room number (\ie, 5, 6, 7, and 8) on a single NVIDIA A100 GPU.
We compute the inference time by averaging 100 samples for each room number.
Our method is trained much faster than the previous state-of-the-art HouseDiffusion method.
For inference time, we observed that 1) in HouseDiffusion, the iterative denoising process is more dominant than the number of rooms and 2) our inference time scales linearly with the number of rooms.

\begin{table}[t]
    \centering
    \begin{adjustbox}{width=0.85\linewidth}
    \begin{tabular}{lcc} \toprule
      Method &  Training Time & Inference Time (5/6/7/8)\\
      \midrule
       HouseGAN++ & 12h & 0.12s / 0.13s / 0.14s / 0.16s\\
       HouseDiffusion & 106h & 10.0s / 10.0s / 10.0s / 10.1s\\
       Ours & 27h & 3.2s / 4.0s / 4.7s / 5.6s  \\
      \bottomrule
    \end{tabular}
    \end{adjustbox}
  \caption{\textbf{Computation cost on a single NVIDIA A100 GPU.}}
  \vspace{-3mm}
  \label{tb:complexity}
\end{table}

\noindent\textbf{Evaluation on doors.}
To assess the alignment quality of doors, we compute GED only on doors by setting the room editing cost to 0.
We use the same generated floorplans as Table~\ref{tb:housediff_graph}.
The results of HouseGAN++, HouseDiffusion, and ours are 1.98, 1.47, and \textbf{1.15}, respectively, showing that our model produces better-aligned doors.

\noindent\textbf{Quantitative gain by constrained decoding.}
We evaluate our model without constrained decoding on graph conditions using the same test set as Table~\ref{tb:housediff_graph}.
We obtain a GED of 1.64 for the variant, where our full model achieves 1.21 while HouseDiffusion achieves 1.55.
The result indicates that constrained decoding is important to achieve better GED. 
This is mainly because, in FML which represents doors as lines, an adjacency is not until that a door completely overlaps on a shared edge between two rooms.

\section{More Related Work}
In addition to the related work referred to in Sec.~\ref{sec:related_work}, we further mention additional studies on floorplan generation.

\noindent\textbf{Completion.}
A prior study~\cite{Hosseini2023floorplan} tackles floorplan generation from panorama images of rooms.  
During training, the model is trained to reconstruct entire floorplans from incomplete floorplans. 
During inference, input panorama images are processed using SfM~\cite{Shabani_2021_ICCV} and converted into partial floorplans. 
After that, partial floorplans are complemented by the trained model.

\noindent\textbf{Room arrangement.}
Our model also can be applied to the room arrangement task formulated as spatial puzzle solving as introduced in PuzzleFusion~\cite{hosseini2023puzzlefusion}.
In this case, each room should be given in a relative coordinate in condition tokens rather than absolute coordinates as boundary conditions.

\noindent\textbf{Text-driven synthesis.}
Tell2Design~\cite{tell2design} generates floorplans from text descriptions with an encoder-decoder architecture.
The core difference between FML and Tell2Design is that Tell2Design conditions floorplans mainly by “natural language”, which sacrifices strictness in exchange for flexibility. 
For example, it would suffer from its ambiguity when distinguishing between two rooms that belong to the same room type. 
FML takes advantage of eliminating such ambiguity by imposing the strict markup-based grammar as defined in Sec.~\ref{sec:fml}.

\section{Failure Cases}
\begin{figure}[h]
  \centering
  \begin{adjustbox}{width=1.0\linewidth}
  \includegraphics{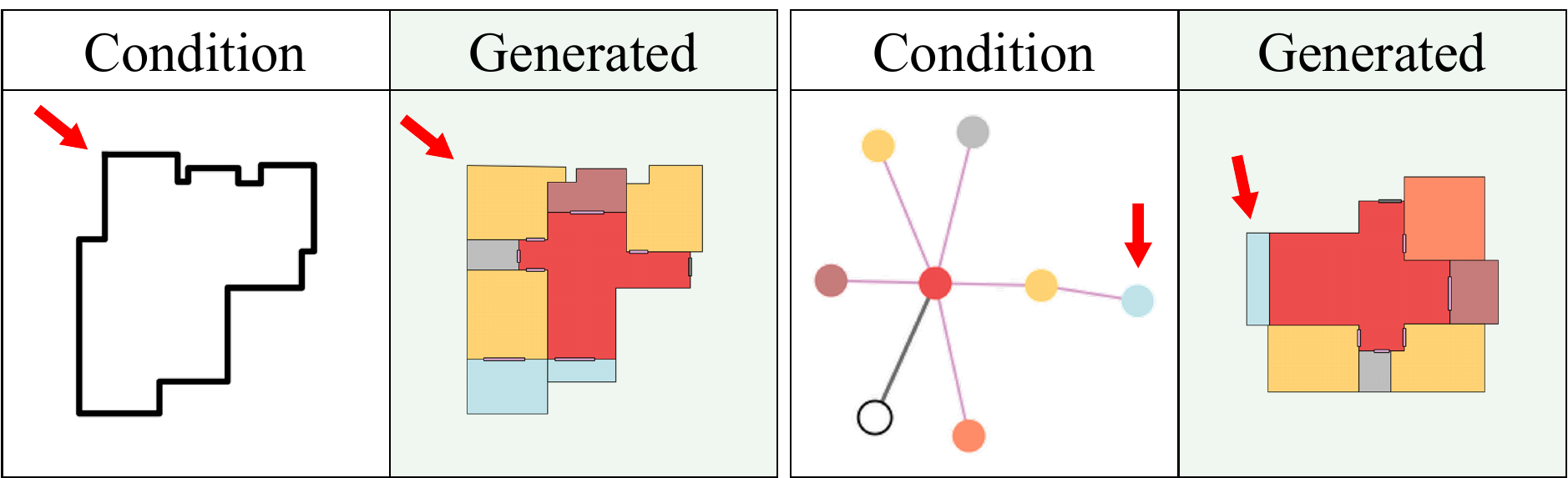}
  \end{adjustbox}
  \caption{\textbf{Failure cases.}}
  \label{fig:failurecases}
\end{figure}

We provide typical failure cases in Fig.~\ref{fig:failurecases}.
It can be observed that 1) our method generates rooms that are not perfectly aligned with boundary conditions and 2) our method places two rooms, that are supposed adjacent, but in distant positions.

\section{More Generated Examples}
\begin{figure}[h]
  \centering
  \begin{adjustbox}{width=1.0\linewidth}
  \includegraphics{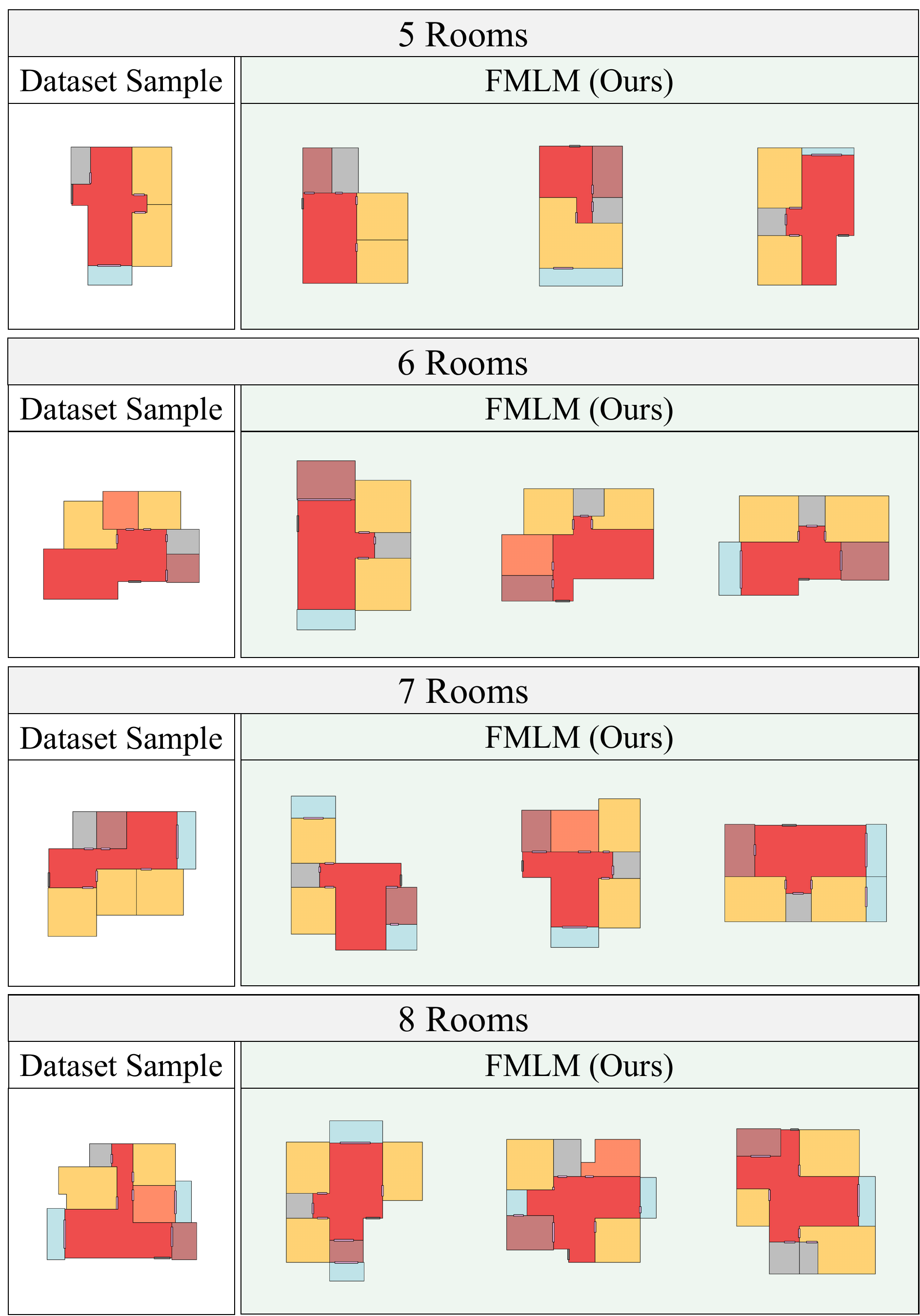}
  \end{adjustbox}
  \caption{\textbf{Number-conditional generation.}}
  \label{fig:qualitative_number}
\end{figure}

\begin{figure*}[t]
  \centering
  \begin{adjustbox}{width=1.0\textwidth}
  \includegraphics{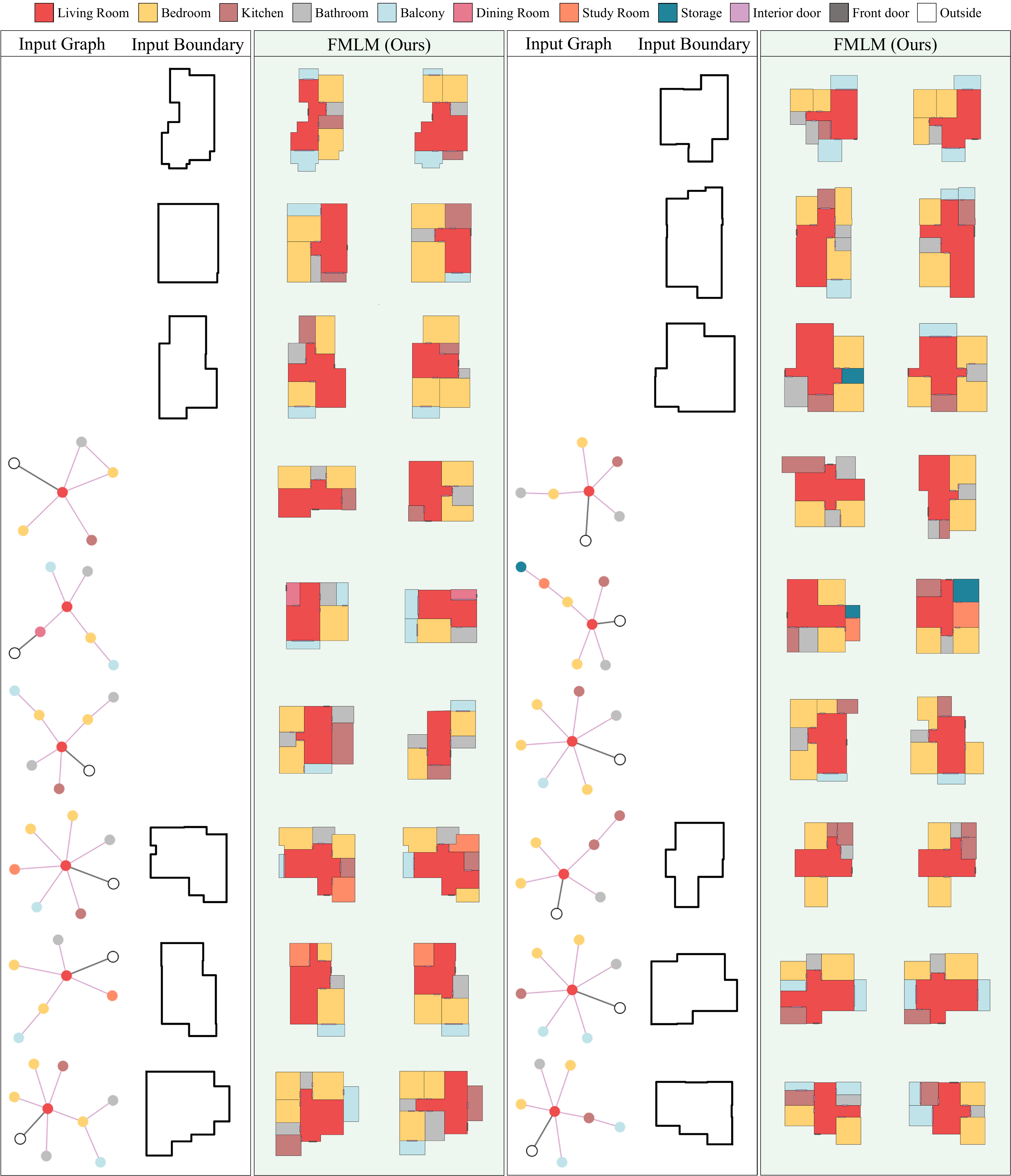}
  \end{adjustbox}
  \caption{\textbf{Additional generated examples.}}
  \label{fig:additional_qualitative}
\end{figure*}

We show the generated floorplans on number-conditional generation in Fig.~\ref{fig:qualitative_number} and additional generated examples on the other conditional generation tasks in Fig.~\ref{fig:additional_qualitative}.

\clearpage

\end{document}